\documentclass[preprints,article,accept,moreauthors,pdftex]{mdpi} 
\firstpage{1} 
\pubvolume{xx}
\issuenum{1}
\articlenumber{5}
\pubyear{2019}
\copyrightyear{2019}
\history{Received: date; Accepted: date; Published: date}
\pdfoutput=1

\Title{Convolutional Neural Networks for Image-Based Corn Kernel Detection and Counting}

\Author{Saeed Khaki $^{1}$*, Hieu Pham $^{2}$, Ye Han $^{2}$, Andy Kuhl $^{2}$, Wade Kent $^{2}$, and Lizhi Wang $^{1}$}

\address{%
$^{1}$ \quad Industrial and Manufacturing Systems Engineering, Iowa State University, Ames, IA, US\\
$^{2}$ \quad Syngenta, Slater, IA, US}

\corres{Correspondence: skhaki@iastate.edu}

\abstract{ Precise  in-season  corn  grain  yield  estimates enable farmers to make real-time accurate harvest and grain marketing  decisions  minimizing  possible  losses  of  profitability. A well developed corn ear can have up to 800 kernels, but manually counting the kernels on an ear of corn is labor-intensive, time consuming and prone to human error. From an algorithmic perspective, the detection of the kernels from a single corn ear image is challenging due to the large number of kernels at different angles and very small distance among the kernels. In this paper, we propose a kernel detection and counting method based on a sliding window approach. The proposed method detects and counts all corn kernels in a single corn ear image taken in uncontrolled lighting conditions. The sliding window approach uses a convolutional neural network (CNN) for kernel detection. Then, a non-maximum suppression (NMS) is applied to remove overlapping detections. Finally, windows that are classified as kernel are passed to another CNN regression model for finding the $(x,y)$ coordinates of the center of kernel image patches. Our experiments indicate that the proposed method can successfully detect the corn kernels with a low detection error and is also able to detect kernels on a batch of corn ears positioned at different angles.}


\begin{document}
%%%%%%%%%%%%%%%%%%%%%%%%%%%%%%%%%%%%%%%%%%

%%%%%%%%%%%%%%%%%%%%%%%%%%%%%%%%%%%%%%%%%%

\section{Introduction}

Commercial corn (Zea mays L.) is processed into numerous food and industrial products and it is widely known as one of the world’s most important grain crops. Corn serves as a source of food for the world and is a key ingredient in both animal feed and the production of bio-fuels. Corn grain yield is driven by optimizing the number of plants per given area and providing sufficient inputs to maximize total kernels per ear within a given environment.

Determining corn grain yield is complicated and requires a detailed understanding of corn breeding, crop physiology, soil fertility, and agronomy. But accurate estimates using simple data inputs can provide reliable information to drive certain management decisions. A well developed corn ear can expect to have over 650-750 kernels. However, various environmental stresses can affect corn ear development impacting the total number of kernels per ear. For instance, drought and heat stress will have a negative correlation with the number of kernels on an ear. Moreover, soil fertility limitations and intense pest pressure throughout a growing season can have adverse effects on total kernels developed resulting in lower total grain yield. Plant breeders work to maximize the amount of material we gain from corn by breeding existing corn with the most resilient, high-yielding genetics. If total kernels per ear, kernel depth, kernel width and estimated kernel weight can be quickly and accurately measured; additional information could be gathered about the crop and allow farmers to make early accurate management decisions.

\subsection{Motivation}
Precise in-season corn grain yield estimates enable farmers to make real-time accurate harvest and grain marketing decisions minimizing possible losses of profitability \cite{zeman2019quantifying}.  These decision can vary from management practices (applying fungicide, nitrogen, fertilizer, etc.) to determining future holding costs with respect to yield futures from the Chicago Mercantile Exchange \cite{shahhosseini2019maize, shi2019, mackenzie2015mechanizing}. Due to the manual labor needed to count the number of kernels on an ear of corn, high-throughput phenotyping is not possible due to the necessary manual labor and the possibility of human error. With modern technology, executing yield estimates in real-time digital applications can be done  efficiently and consistently, compared to past methods, while providing the ability to make historical comparisons following harvest \cite{ziamtsov2019machine}.  Agronomically, accurate in-season yield estimates deliver the unique potential for agronomists and farmers to diagnose potential issues that have or may impact corn grain yield, and equips them with the informed knowledge to make real-time decision with respect to their harvest. Recently, image-processing, machine learning, and deep learning have shown great potential in progressing the digital capabilities needed for the future of agriculture. These techniques have shown to be reliably in high-throughput phenotyping and in enable farmers to make real-time decision, something that was previously not possible.

Due to the need to count corn kernels on numerous ears and because of the manual limitation of this task, this work proposes a new deep learning approach to estimating the number of kernels on an ear of corn that can be used for real-time decision making. This methodology takes an image of a single or multiple ears of corn and outputs the estimated number of kernels in the entire image with no assumptions on either the background environment nor the lighting conditions of the image.
%-------------------------------------------------------------------------
\subsection{Literature Review and Related Works}

Succinctly, machine learning is a method of data analysis to automatically identify patterns within data which can be tabular, images, text, etc. The process of machine learning requires building a model on an initial dataset, called the training dataset, and then using an independent dataset, called the test set, to validate the perform of the model on data which was not used for training. This procedure allows for a true representation of the accuracy of the trained machine learning model. There exists a large literature on various machine learning models in a variety of domains \cite{mohri2018foundations, domingos2012few, libbrecht2015machine, sun2013survey}. However, we will not provide a review here as ultimately we want to focus our attention on a special case of machine learning often referred to as deep learning.

%\textcolor{red}{\textbf{GENERAL DEEP LEARNING AND OBJECT DETECTION REVIEW HERE.}}

Deep learning models are representation learning methods with multiple levels of representations. Each level of representations has nonlinear modules to transform the representation at the current level (starting with the raw input) to a slightly more abstract level \cite{lecun2015deep}. Deep neural networks also belong to a class of universal approximators \cite{hornik1989multilayer}, which means regardless of what function we want to learn, they can be used to represent such function \cite{goodfellow2016deep}. Deep learning models automatically perform feature extraction on input data without the need of using any handcrafted input of features.

As one of the fundamental component of computer vision, object detection provides information about the concepts and locations of
objects contained in each image \cite{zhao2019object}. As such, the goal of object detection is to localize objects in a given image and determine which category each object belongs to. Traditional object detection methods first extract feature descriptors such as HOG \cite{dalal2005histograms} and SIFT \cite{lowe2004distinctive}. Then, they train a classifier such as a support vector machine (SVM) \cite{cortes1995support} and AdaBoost \cite{freund1995desicion} based on extracted feature descriptors to distinguish a target object from all the other categories. More recently, deep learning based object detection methods have been proposed. These methods such as single shot detection (SSD) \cite{liu2016ssd}, you only look once (YOLO) \cite{redmon2016you}, and fast R-CNN \cite{girshick2015fast} automatically extract necessary feature descriptors which significantly improves their accuracies compared to traditional object detection methods. However, these methods are very data hungry and computationally expensive to train.

In terms of applying machine learning, image processing, and deep learning for object detection in agriculture, there has been no shortage of use-cases. Traditional image processing based approaches often referred to as image segmentation (filtering, watershedding, thresholding, etc.) have been applied to mangoes, apples, tomatoes, and grapes for detecting and counting within images. \citep{pal1993review, Yamamoto2013, SENGUPTA201451, Zhang, Qureshi2017, Gnadinger2017}. Although, successful, these approaches typically require large amounts of high-resolution images with minimal noise, cannot handle large variation in crop sizes, and can only identify a single crop per image. 

Using a machine learning approach, Ok et al. \cite{ok2012evaluation} demonstrated that the random forest (RF) algorithm \cite{breiman2001random} and maximum likelihood classification \cite{bolstad1991rapid} were indeed suitable at successfully classifying wheat, rice, corn, sugar beet, tomatoes, and peppers within fields using satellite imagery. Additionally, Zawbaa et al. \cite{zawbaa2014automatic} designed an experiment to automatically classify images of apples, strawberries, and oranges using RF and $k$- nearest neighbors model \cite{cover1967nearest}. Their study further demonstrates the success that machine learning capabilities have in agriculture.  Moreover, Guo et al. \cite{Guo2018} applied a quadratic-SVM \cite{cortes1995support} to accurately detect and count sorghum heads from unmanned aerial vehicle (drone) images. Although these example show the power that modern machine learning has in object detection, specifically in agriculture, they are not without fault. Namely, tradition machine learning approaches cannot generalize well to objects with varying image resolutions, different image scaling (distance from camera to object) and different object orientations (object angles). 

Due to the power of deep learning being able to recognize multiple objects within images and the lack of requirements towards object orientations, there has been a large amount of recent literature in deep learning in agriculture. In 2019, Ghosal et al. applied their method based on a RetinaNet to detect and count sorghum heads from drone images \cite{ghosal2019weakly}.  This deep learning approach significantly out performed prior sorghum detection and counting work by Guo et al. \cite{Guo2018}. Various other deep learning models have also been proposed in disease detection, quality assessment and detection and counting of various crops \cite{da2020computer,kuricheti2019computer,dhingra2019novel, agarwal2019computer}. DeepCrop is an image repository consisting of 31,147 images with over 49,000 annotations from 31 different crop classes \cite{jin2020hybrid}. This dataset has been instrumental in the advancement of object detection in agriculture where often times gathering annotated data is a challenge \cite{xie2019deep, joseph2019harvestable}. With the advent of transfer learning, models can be pre-trained on such datasets and have their information transferred to detect similar objects without the need for long training times \cite{tan2018survey}. Due to the large literature combining deep learning and agriculture, we cannot do justice in providing a comprehensive review. Instead, we point the reader towards a survey paper which gives a thorough overview of image-based plant phenotyping using deep learning \cite{Jiang2020}.

We have provided an overview of image processing, machine learning, and deep learning in various agricultural tasks, but now we turn out attention to the focus of this paper, namely, work that has been completed in counting corn kernels. In 2014, Zhao et al.  \cite{zhao2014automatic} applied traditional image processing based approaches to count kernels, but was still limited to the previously mention limitations of requiring high resolution images, low signal to noise ratio, and only being able to count from a single ear per image. Grift et al. \cite{grift2017semi} also invoked an image processing based approach but limits ear images to be taken within a soft box fitted with controlled and uniform lighting conditions. Moreover, the images in their study contained 360 degree photos, that is, they designed a special lighting box so that lighting conditions were controlled and to take complete photos of the ear. Ni et al. in 2018 \cite{ni2018convolution} and Li et al. in 2019 \cite{li2019corn} both utilized deep learning to count corn kernels, however, their algorithms were designed to count kernels  already removed from the cob. Although both were able to accurately count kernels, their problem is easier than directly counting kernels while on the ear, due to the distinct spacing between kernels in their images. Additionally, this process does not allow for real-time in-field decision making due to having to shell the kernels off the ear before proceeding with the counting. Although, each of these previous methods have ``moved the needle'' in regards to kernel counting there is not a concise method which address all of theses limitations.

Due to the difficult nature of this problem and the demand for in-field corn kernel count estimates, we propose a deep learning approach to detect and count corn kernels where kernels are still intact on an ear simply using a 180 degree image. This approach will be robust enough to handle any set of ears regardless of the orient of the ears and the light conditions present.

%\section{Methodology}

\section{Methodology}\label{sec:method}

\begin{figure}[H]
    \centering
    \includegraphics[scale=0.04]{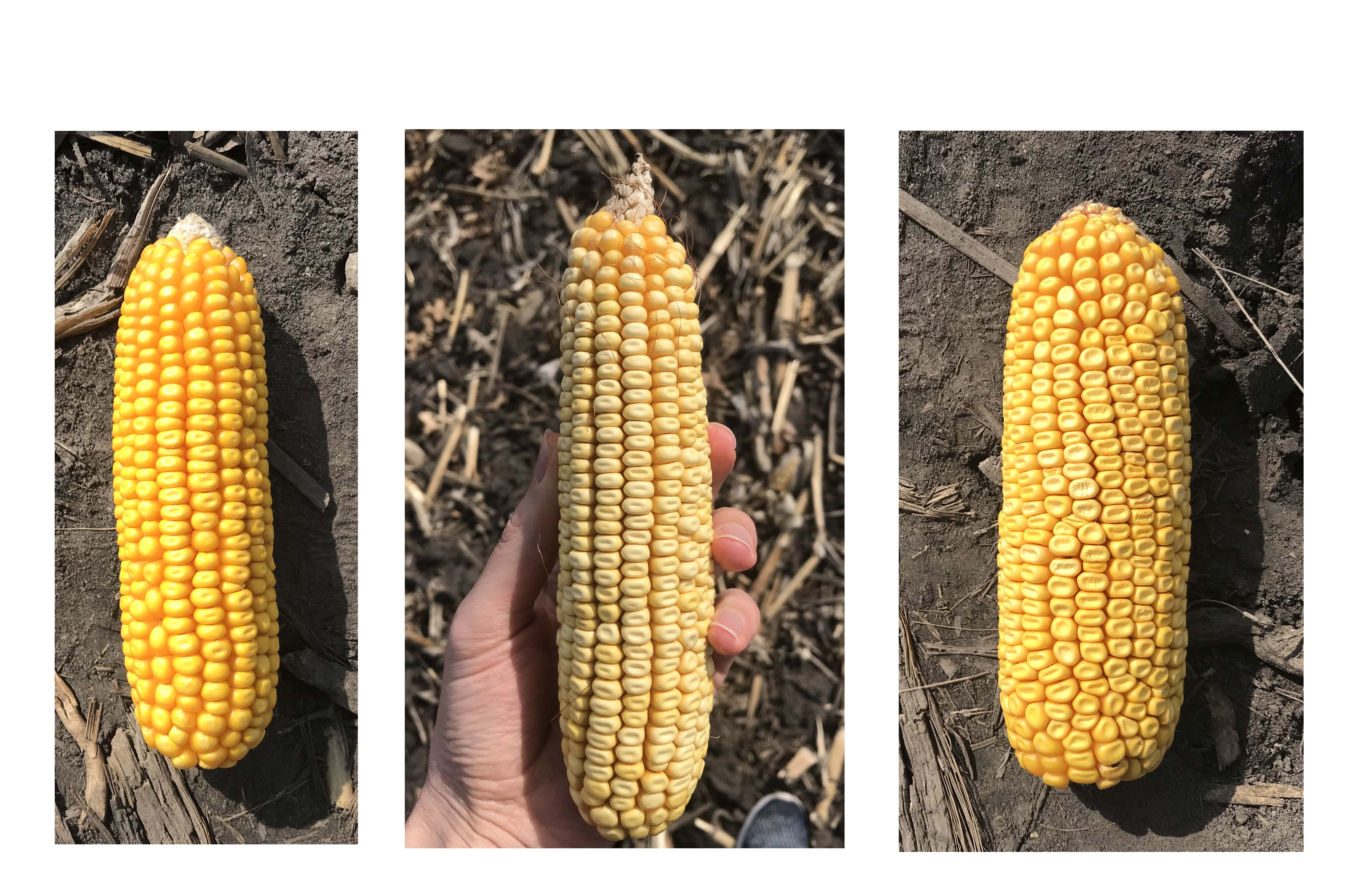}
    \caption{Three different corn ears.}
    \label{fig:corn_ears}
\end{figure}

The goal of this study is to localize and count corn kernels in a corn ear image taken in uncontrolled lighting conditions. To solve this problem, we first detect all kernels in a corn ear image and then estimate the total number of kernels by counting the number of detected kernels. As a result, the underlying research problem is a single class object detection problem.  As shown in Figure \ref{fig:corn_ears}, the number of objects (kernels) in a corn ear is extensive (up to 800 kernels) and the objects are in close proximity to one another, making the problem more challenging.

\begin{figure}[H]
    \centering
    \includegraphics[scale=0.10]{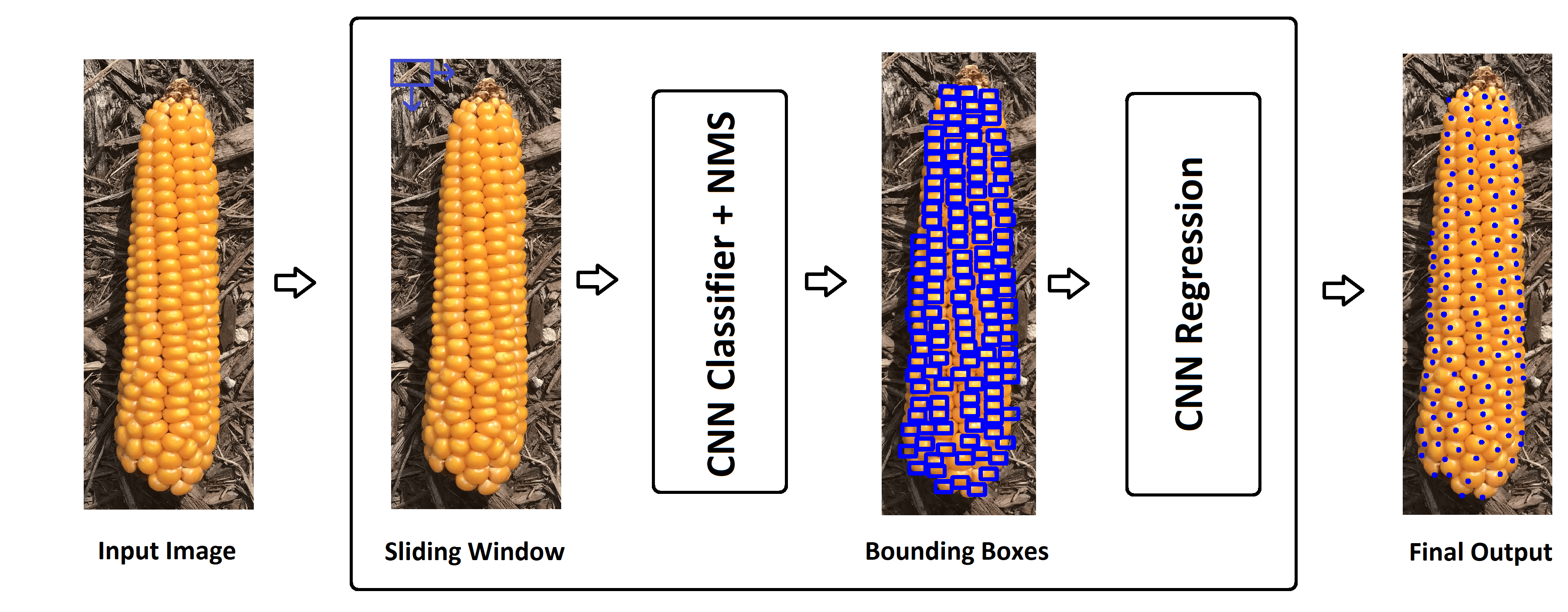}
    \caption{Modeling structure of our proposed corn kernel detection method. A detailed description is given in Section \ref{sec:method}.}
    \label{fig:modeling_str}
\end{figure}

We use a sliding window approach for kernel detection in this study. At each window position, a convolutional neural network classifier returns a confidence value representing its certainty that the current window contains a kernel or not. After computing all confidence values, a NMS is applied to remove redundant and overlapping detections. Finally, windows that are classified as a kernel are passed to a regression model. The regression model predicts $(x,y)$ coordinates of the center of kernels given image patch of kernels. Figure \ref{fig:modeling_str} shows the modeling structure of our proposed corn kernel detection method. Detailed description of the kernel classifier and the regression model is provided in the following sections. In this study, we did not use popular object detection methods such as SSD \cite{liu2016ssd}, YOLO \cite{redmon2016you}, and fast R-CNN \cite{girshick2015fast} mainly because these methods need considerable amount of annotated images which do not exist publicly for the corn kernel detection. In addition, we could not use transfer learning since corn kernel detection is very different than other object detection tasks such as leaf or human detections.

\subsection{Corn Kernel Classifier}\label{sec:classifier}

In this paper, we apply a sliding window approach for kernel detection problem which requires a supervised learning model to classify the current window as either kernel or non-kernel. We use a CNN to classify image patches as CNNs have been shown to be a very powerful method for the image classification task \cite{krizhevsky2012imagenet,he2016deep,szegedy2015going}. The CNN model takes in image patches with size of $32 \times 32$ pixels. The CNN architecture for kernel classification is defined in Table \ref{tab:CNN_kernel_classifier}. All layers are followed by a batch normalization \cite{ioffe2015batch} and ReLU nonlinearity except the final fully connected layer which has a sigmoid activation function to produce a confidence value representing the CNN's certainty that an input image patch contains a kernel or not. Down sampling is performed with average pooling layers. We do not use dropout \cite{srivastava2014dropout}, following the practice in \cite{ioffe2015batch}.

\begin{table}[h]
 \caption{The CNN architecture for kernel classification.}
    \label{tab:CNN_kernel_classifier}
    \centering
    \begin{tabular}{c|c|c|c}
    \hline
    \hline
       Type / Stride  & Filter Size & \# of Filters & Output Size\\
       \hline
        Conv/s1 &$3\times3$&  32& $30 \times 30 \times 32$\\
        \hline
        Conv/s1 &$3\times3$&  32&$28 \times 28 \times 32$\\
        \hline
        Avg pool/s2 &$2\times2$&  - &$14 \times 14 \times 32$\\
        \hline
        Conv/s1 &$3\times3$&  64 &$12 \times 12 \times 64$\\
        \hline
        Conv/s1 &$3\times3$&  64 &$10 \times 10 \times 64$\\
        \hline
         Conv/s1 &$3\times3$&  64 &$8 \times 8 \times 64$\\
        \hline
         Avg pool/s1 &$7\times7$&  - &$2 \times 2 \times 64$\\
        \hline
         \multicolumn{4}{c}{FC-256}\\
         \hline
         \multicolumn{4}{c}{FC-128}\\
         \hline
         \multicolumn{4}{c}{Sigmoid}\\
        \hline
        
    \end{tabular}
   
\end{table}

\subsection{Regression Model}\label{sec:reg}

As shown in  Figure \ref{fig:corn_ears}, the kernels are very close to each other on corn ears. As such, if we visualized all detected kernels with bounding boxes in a corn ear image, it would be almost impossible to see the corn ear, especially on the left and right sides of the ear due to having many close bounding boxes. Furthermore, some kernels have different shapes and angles which might not fit perfectly in a rectangle bounding boxes. As such, we use a convolutional neural network as a regression model which takes in an image of kernel with size of $32 \times 32$ pixels and predicts $(x,y)$ coordinates of the center of the kernel. The primary reason for not simply using the center of the windows being classified as kernel as the center of detected kernels is that the center of the kernels are not always in the center of the windows, especially for the kernels on the sides of the corn ear. The CNN architecture for finding the $(x,y)$ coordinates of the center of kernel image is defined in Table \ref{tab:CNN_regression}. All layers are followed by ReLU nonlinearity except the final fully connected layer which has no nonlinearity. Down sampling is performed with max pooling layers. We did not use dropout for this model as it did not improve overall performance. The regression model is applied only on the final windows being classified as a kernel after the NMS. As such, the proposed regression model does not add a lot of computational cost to the kernel detection approach considering the number of final windows being classified as kernel is small.
%approximately between 100 and 400 depending on the size of corn ears.

\begin{table}[h]
 \caption{The CNN architecture for finding the $(x,y)$ coordinates of the center of a kernel image.}
    \label{tab:CNN_regression}
    \centering
    \begin{tabular}{c|c|c|c}
    \hline
    \hline
       Type / Stride  & Filter Size & \# of Filters & Output Size\\
       \hline
        Conv/s1 &$3\times3$&  32& $30 \times 30 \times 32$\\
        \hline
        Conv/s1 &$3\times3$&  32 &$28 \times 28 \times 32$\\
        \hline
        Max pool/s2 &$2\times2$&  - &$14 \times 14 \times 32$\\
        \hline
        Conv/s1 &$3\times3$&  64 &$12 \times 12 \times 64$\\
        \hline
        Conv/s1 &$3\times3$&  64 &$10 \times 10 \times 64$\\
        \hline
         Conv/s1 &$3\times3$&  64 &$8 \times 8 \times 64$\\
        \hline
         Max pool/s2 &$2\times2$&  - &$4 \times 4 \times 64$\\
        \hline
         \multicolumn{4}{c}{FC-100}\\
         \hline
         \multicolumn{4}{c}{FC-50}\\
         \hline
         \multicolumn{4}{c}{FC-10}\\
         \hline
         \multicolumn{4}{c}{FC-2}\\
        \hline
        
    \end{tabular}
   
\end{table}

\section{Experiments and Results}

This section presents the dataset used for our experiments, the training hyperparameters, and the final results. We consider standard evaluation measures such as false positive (FP), false negative (FN), accuracy, and f-score. All our experiments were conducted in Python using the TensorFlow \cite{abadi2016tensorflow} library on a NVIDIA Tesla V100 GPU.

\subsection{Dataset}

\begin{figure}[H]
    \centering
    \includegraphics[scale=0.35]{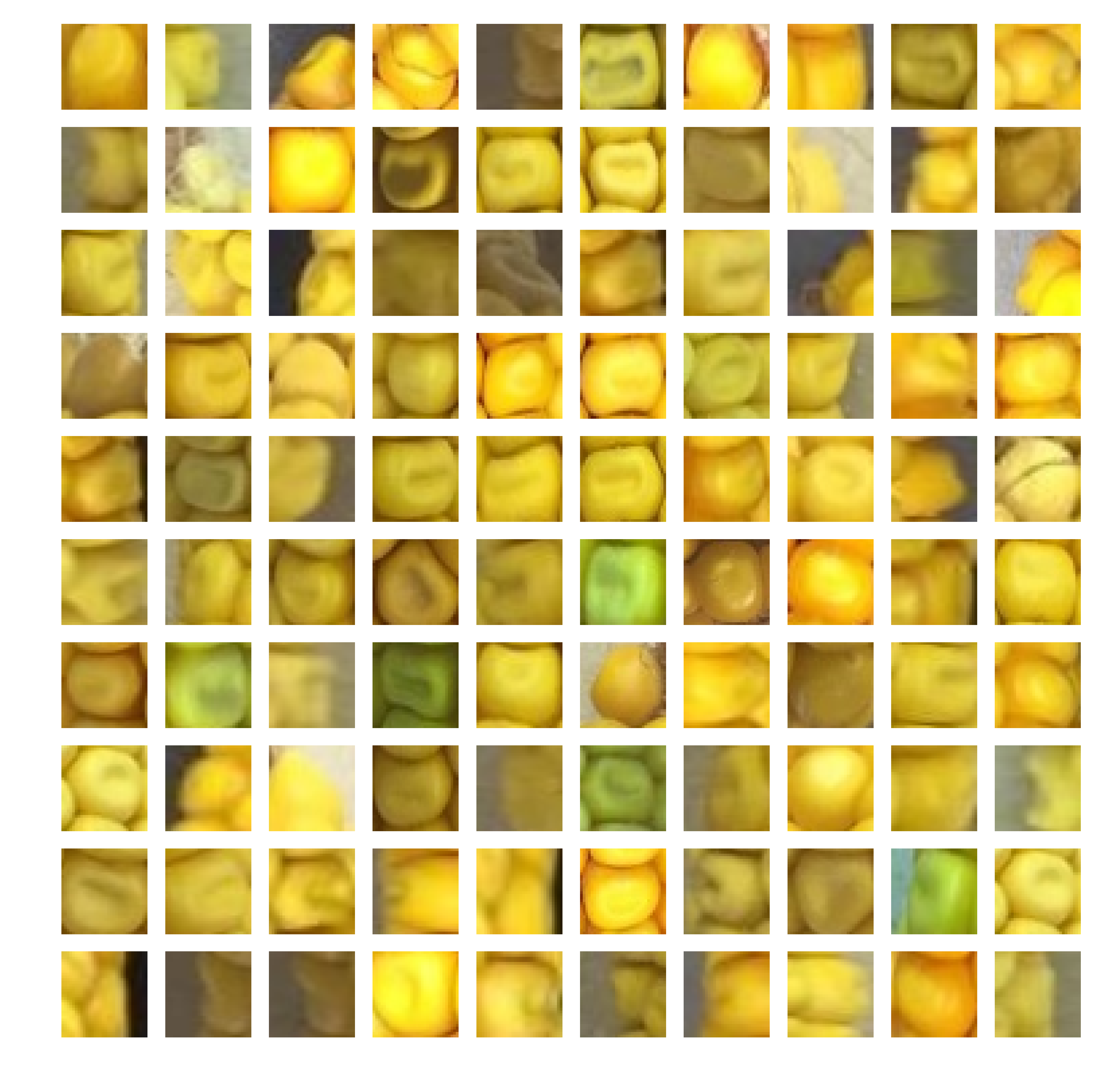}
    \caption{A random subset of kernel images.}
    \label{fig:kernel}
\end{figure}

\begin{figure}[H]
    \centering
    \includegraphics[scale=0.35]{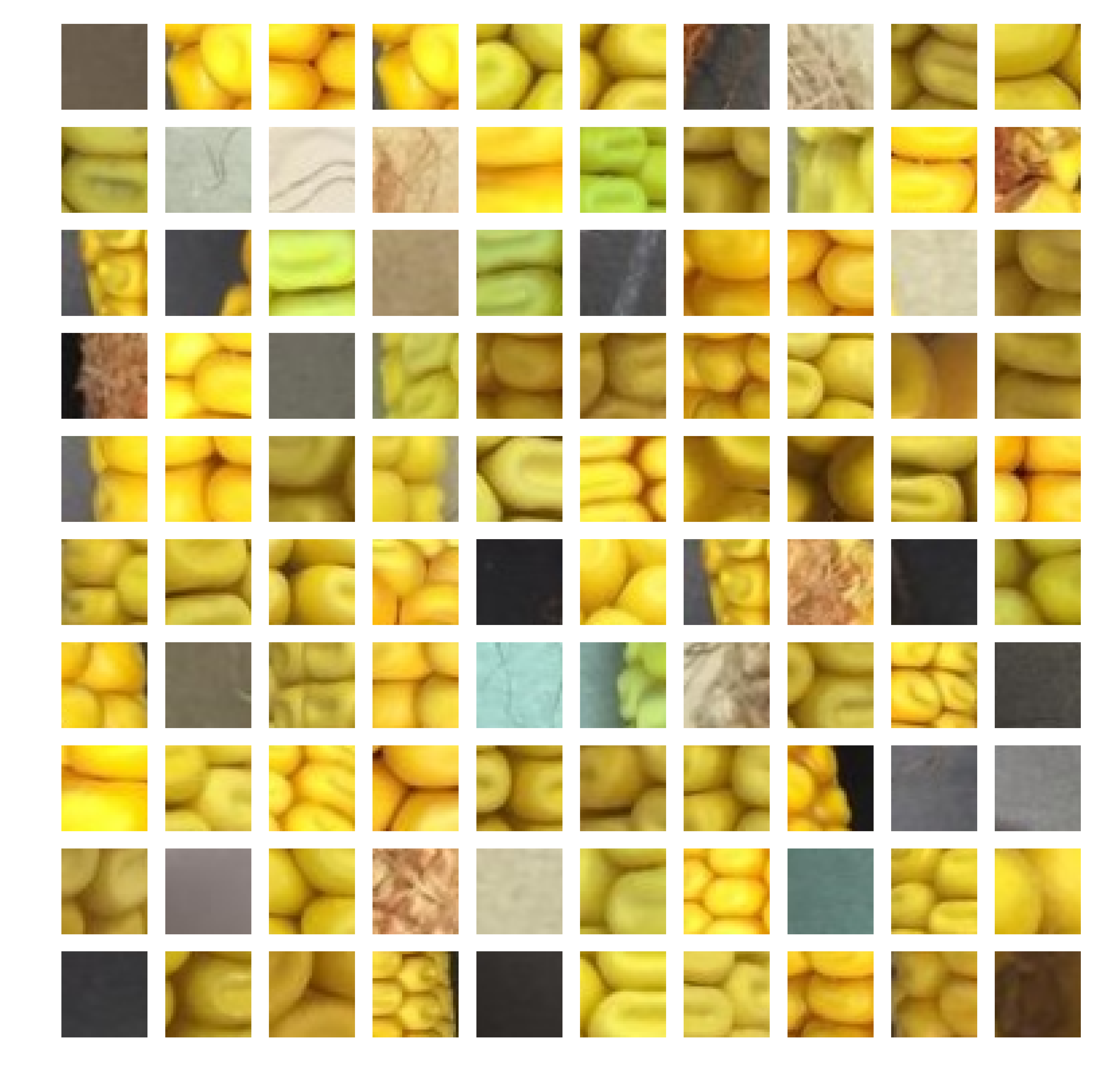}
    \caption{A random subset of non-kernel images.}
    \label{fig:nonkernel}
\end{figure}

\begin{figure}[H]
    \centering
    \includegraphics[scale=0.35]{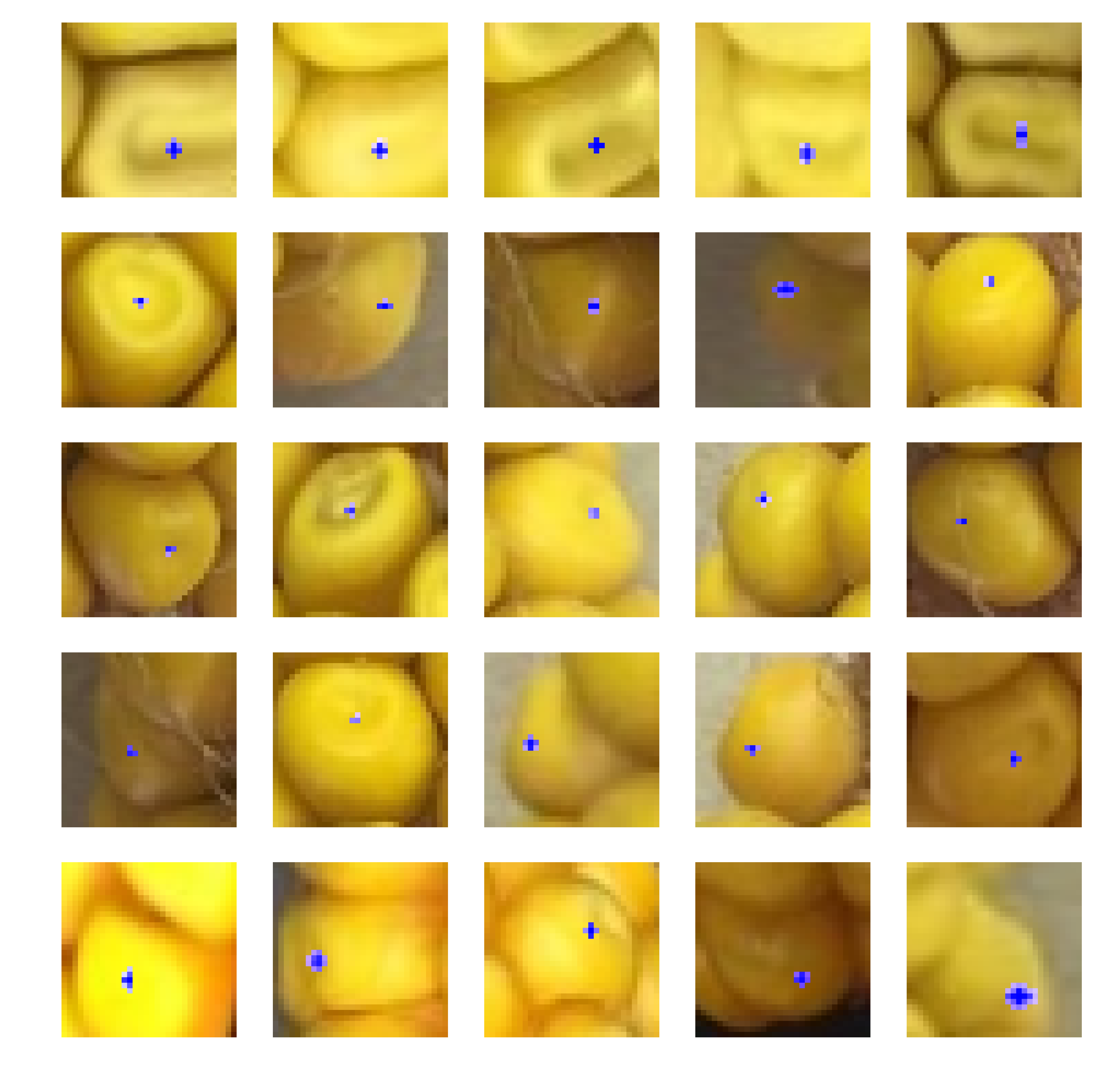}
    \caption{A random subset of annotated kernel images. The blue dot indicates the center of the kernel.}
    \label{fig:center}
\end{figure}

The proposed sliding window approach requires a trained kernel classifier before it can be applied. Therefore, positive samples of kernels and negative samples of non-kernel are necessary. The authors manually cut and labeled kernel and non-kernel images from 43 different corn ear images to generate the training dataset. Each kernel sample is cut out and scaled to $32\times32$ pixels. Negative samples are generated in the same way using random crops at different positions. The positive samples only include image of one kernel. If the image patch contains two or more kernels, it is considered a negative sample. The training dataset consists of 6,978 kernel and 9,413 non-kernel samples. Figure \ref{fig:kernel} and \ref{fig:nonkernel} show a subset of kernel and non-kernel images, respectively. For the regression model, we only used the kernel image part of the dataset. We manually labeled the kernel images by finding the $(x,y)$ coordinates of their centers using Labelme \cite{labelme2016} software. Figure \ref{fig:center} depicts a subset of annotated kernel images.

%We randomly took 5\% of the dataset as validation data and used the rest as the training data. We augmented the 

\subsection{Corn Kernel Classifier Training}
\begin{figure}[H]
    \centering
    \includegraphics[scale=0.28]{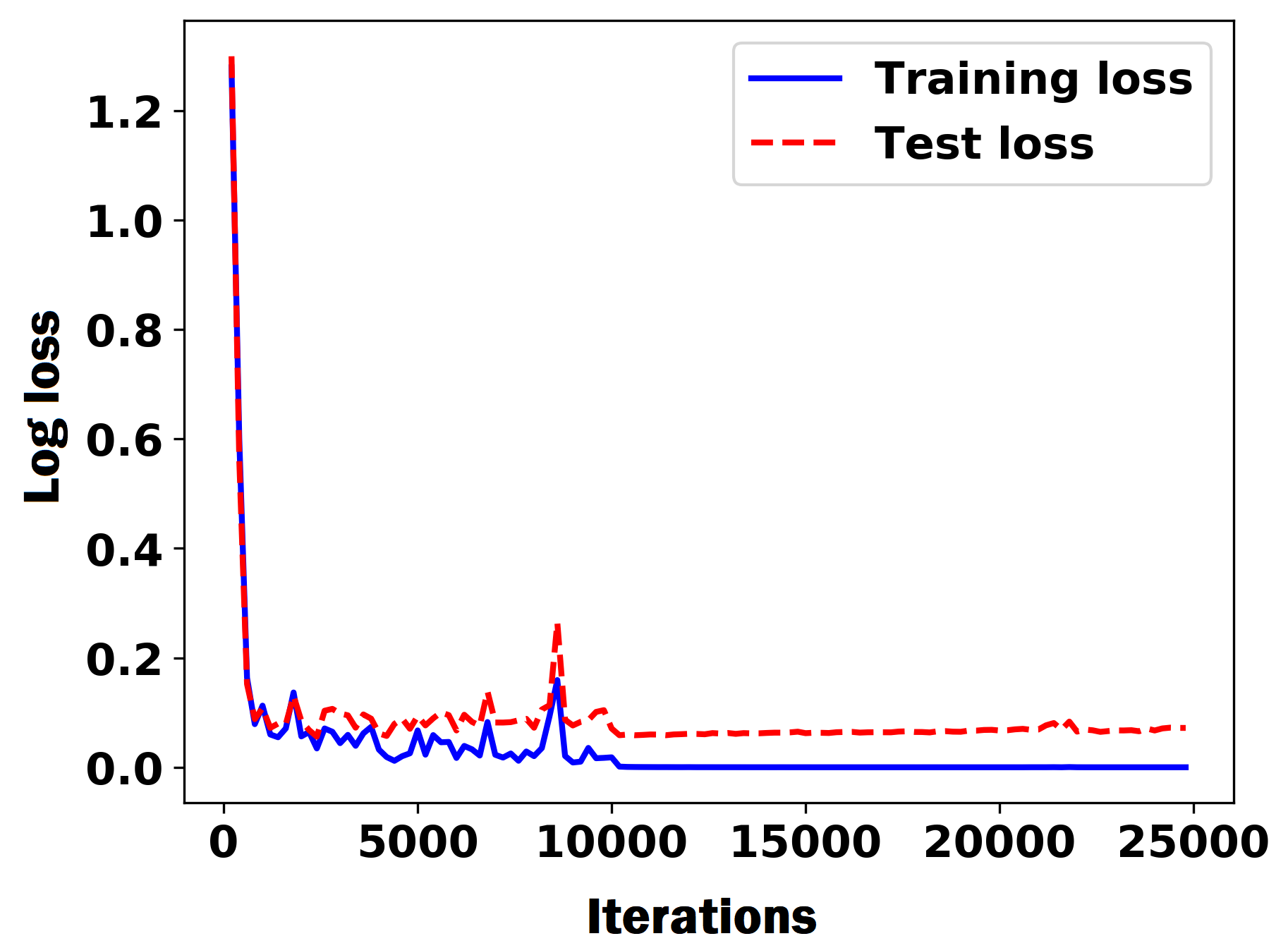}
    \caption{Plot of the log loss of the CNN classifier during training process.}
    \label{fig:loss}
\end{figure}

We trained the CNN as described in section \ref{sec:classifier} for kernel classification using the following training hyperparameters. The weights were initialized with the Xavier initialization \cite{glorot2010understanding}. A stochastic gradient descent (SGD) was used with a mini-batch size of 128. The learning rate started from 0.03\% and was reduced to 0.01\% when error plateaued. The model was trained for 25,000 iterations.  Adam optimizer \cite{kingma2014adam} was used to minimize the log loss. For our data, we randomly took 20\% of the data as the test data (3,278 images) and used the rest as the training data. We augmented around 70\% the training data with flip and color augmentations. After augmentation, we had total of 22,292 training images. Figure \ref{fig:loss} shows the plot of training and test losses for the CNN. To better evaluate the CNN classifier, a comparison of the CNN classifier with the HOG+SVM model was performed \cite{dalal2005histograms}. This model uses the Histogram of Oriented Gradient (HOG) to extract edge features to describe the object’s shape and then trains a support vector machine (SVM) classifier based on the extracted features. The best results achieved for the HOG+SVM were with the parameters $4\times4$ pixels per cell, 2 cells per block, and 9 histogram bins. Table \ref{tab:Classifers_results} compares the performances of the CNN and HOG+SVM classifiers on the training and test datasets. We used the CNN model as our final kernel classifier because it resulted in a more reliable kernel detection and counting. Moreover, the CNN model can successfully generalize the prediction to different backgrounds.  

\begin{table}[H]
 \caption{Performance comparison of the CNN and HOG+SVM classifiers on the training and test datasets.}\label{tab:Classifers_results}
    \centering
    \renewcommand{\arraystretch}{1.8}
    \begin{tabular}{|c|c|c|c|c|c|}
    \cline{2-6}
   \multicolumn{1}{c|}{}   &  \multirow{2}{*}{Classifier} &  \multicolumn{4}{c|}{Evaluation Measures} \\
       \cline{3-6}
    \multicolumn{1}{c|}{}   & &  FP & FN & Accuracy & F-score\\
        \hline

    \multirow{2}{*}{\rotatebox{90}{Training}}  &  HOG+SVM & 596 & 595& 0.947 & 0.937\\
      \cline{2-6}
     &  CNN & 0 & 0 & 1.0 & 1.0\\
       \hline  \hline
       
         \multirow{2}{*}{\rotatebox{90}{Test}} &  HOG+SVM & 135&135& 0.918&0.906\\
       \cline{2-6}
     &  CNN & 19 & 22& 0.987 & 0.985\\
     \hline

    \end{tabular}
   
\end{table}

Table \ref{tab:Classifers_results} indicates that the CNN model outperforms the HOG+SVM model with respect to all evaluation measures. One of the reasons for the higher accuracy of the CNN classifier compared to the HOG+SVM is that the CNN automatically extracts necessary features from the data. However, the HOG+SVM model is faster to train and test from computational perspective.

\subsection{Regression Model Training}

\begin{figure}[H]
    \centering
    \includegraphics[scale=0.135]{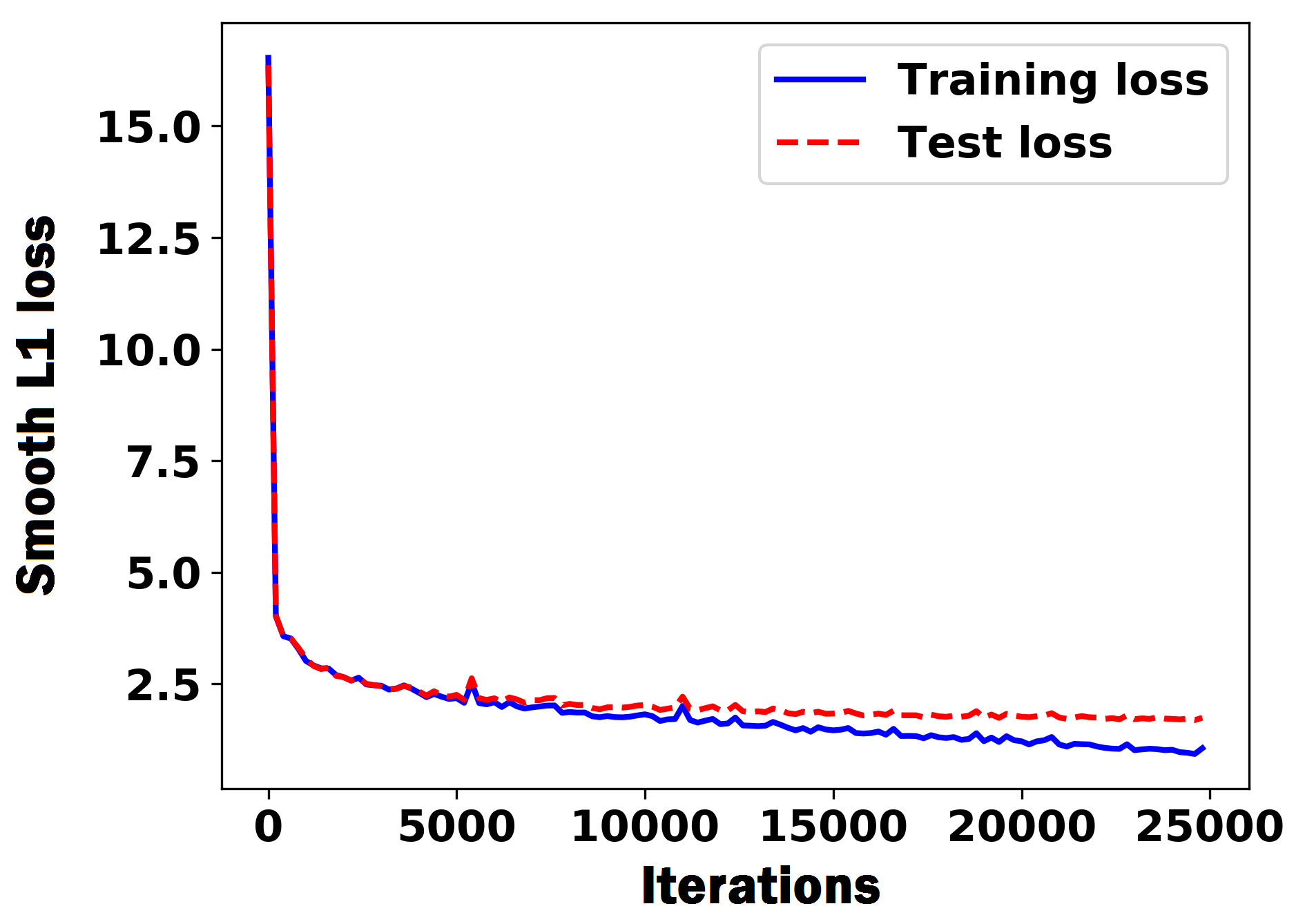}
    \caption{Plot of the smooth $\textit{L}_1$ loss of the CNN regression model during training process.}
    \label{fig:loss_reg}
\end{figure}

The CNN model was trained as described in section \ref{sec:reg} for finding the $(x,y)$ coordinates of the center of a kernel image using the following training hyperparameters. The weights were initialized with the Xavier initialization. A stochastic gradient descent (SGD) was utilized with a mini-batch size of 45. The model was trained for 25,000 iterations with the learning rate of 0.03\%.  Adam optimizer was used to minimize the smooth $\textit{L}_1$ loss as in \cite{girshick2015fast}, which is less sensitive to the outliers compared to the $\textit{L}_2$ loss. We randomly took 20\% of the data as the test data (1,396 images) and used the rest as the training data (5,582 images). Figure \ref{fig:loss_reg} shows the plot of the training and test losses for the CNN regression model.

\subsection{Final Results}

Having trained our kernel detection model, we can now apply the sliding window approach with the trained CNN classifier on several test images containing full ears. After applying the NMS, the windows that were classified as kernel were passed to the regression model for finding their corresponding centers. We used window size of $32\times22$ for the sliding window approach. To fully evaluate the proposed approach, we tested the approach on the multiple corn ears with different angles, backgrounds and lighting conditions. Farmers and agronomists assume that corn ears are symmetric \cite{bennetzen2008handbook}. As such, they count the number of kernels on the one side and then double it to approximately find the total number of corn kernels on a corn ear. We used a similar approach except that we multiplied the number of detected kernels on the one side by 2.5 because around 2 columns of kernels on the very left and right sides of the ear are not captured in the image and consequently not counted. The inference time for a corn ear is 5.79 seconds.

\begin{figure}[H]
    \centering
   \includegraphics[scale=0.22]{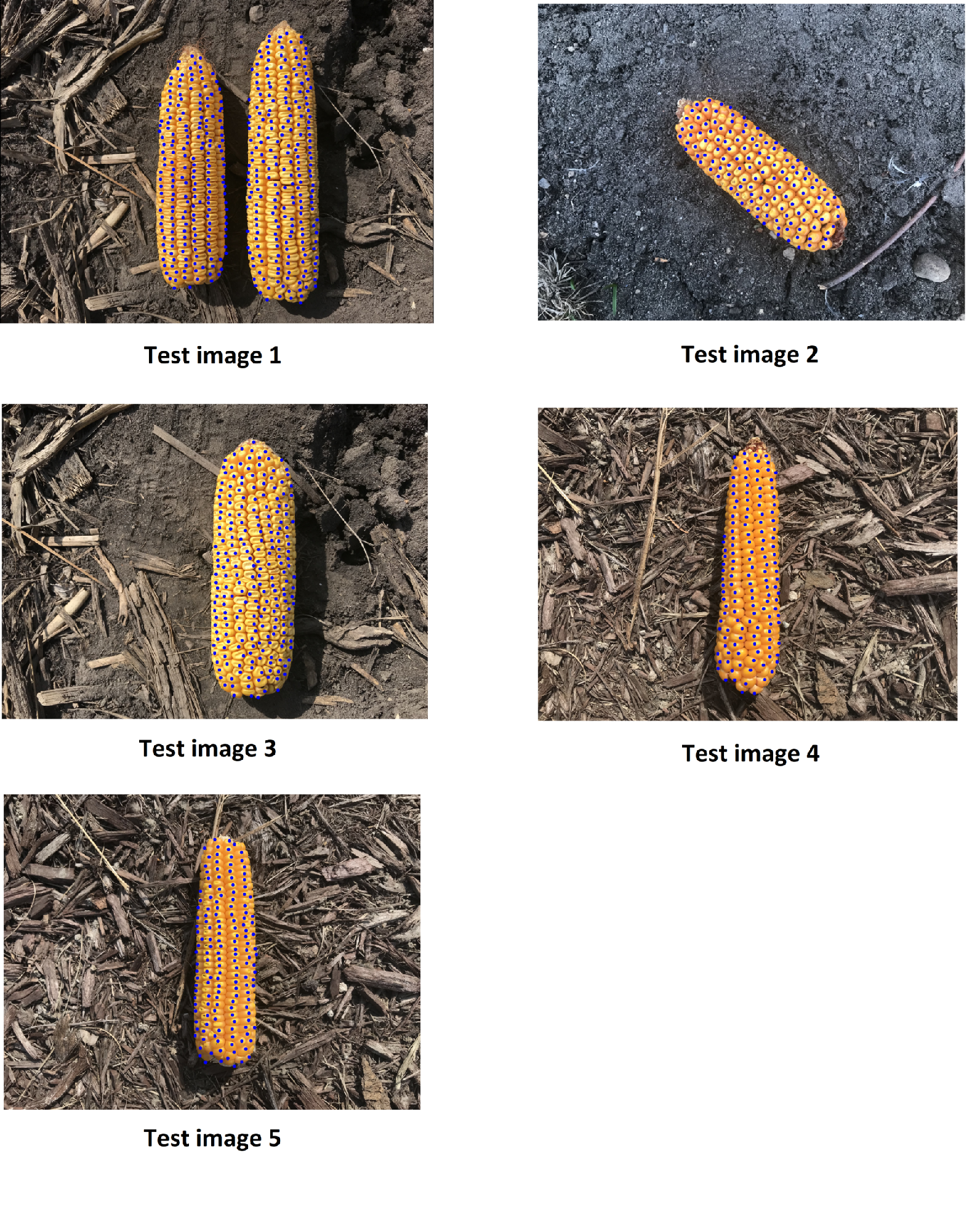}
    \caption{The results of the proposed approach on 5 different test images.}
    \label{fig:result_whole}
\end{figure}

Figure \ref{fig:result_whole} shows the results of the proposed approach on 5 different test images. As shown in Figure \ref{fig:result_whole}, the proposed approach successfully found the most of kernels in the test image 1. Test image 2 shows the results of the proposed approach on the image of an angled corn ear. This image is considered a difficult test image because we did not include any angled kernel image in the training dataset. But, the results indicate that the approach can generalize the detection to the images of angled corn ears. We also applied the approach on another difficult test image of a corn ear whose kernels are slightly angled, and as shown in test image 3, the proposed approach is still able to detect most of the kernels. Test images 4 and 5 also show the performance of the proposed method on two other test corn ears. Table \ref{tab:descrip} shows the predicted and the ground truth numbers of the
kernels on test images shown in Figure \ref{fig:result_whole}. Our proposed approach has the following advantages for kernel counting: (1) our proposed approach can be used on a batch of corn ears, and (2) our proposed approach can be used on a slightly angled corn ear.

\begin{table}[H]
 
    \caption{The predicted and the ground truth numbers of the
kernels on test images shown in Figure \ref{fig:result_whole}}
    \label{tab:descrip}
    \centering
    \begin{tabular}{|c|c|c|}
    \hline
         Test Image & \begin{tabular}{c}
            Predicted \\Number of Kernels
         \end{tabular} & \begin{tabular}{c}
            Actual \\Numbers of Kernels
         \end{tabular}  \\
    \hline 
    1 &   1,012 & 1,046\\ 
         \hline
         
    2  &   312&323\\
    \hline
     3 &   550&585\\
         \hline
         
    4 &  342& 296\\
         
    \hline
    
     5 &  390& 394\\
         
        \hline

    \end{tabular}

\end{table}

To completely evaluate our proposed approach, we manually counted the entire number of kernels on 20 genetically different corn ears and used the proposed method to estimate the number of kernels on these corn ears. We also implemented the method proposed by Chuan et al. \cite{wang2015deep} called Deep Crowd which was originally developed for people counting in extremely dense crowds using convolutional neural networks. Deep Crowd is one of the state-of-
the-art methods proposed for people counting in dense crowds in the literature. The people counting in extremely dense crowds problem is similar to the corn kernel counting problem for two main reasons: (1) they both want to count a large number of objects, and (2) objects are very close to each other. We used the following hyperparameters for training the Deep Crowd method. We used the exact same network architecture as in \cite{wang2015deep}. We used 43 corn ear images with $768\times1024$ pixels as training data. We randomly cropped 120 patches with $227\times 227$ pixels from each ear image which resulted in the 5,160 patches for training the CNN. We also augmented the training data using color and flip augmentations. The CNN was trained using SGD with learning rate of 0.03\%. 

%since we want to count the number of objects in both problems which
%We compared our proposed method with this method because the  

\begin{figure}[H]
    \centering
    \includegraphics[scale=0.30]{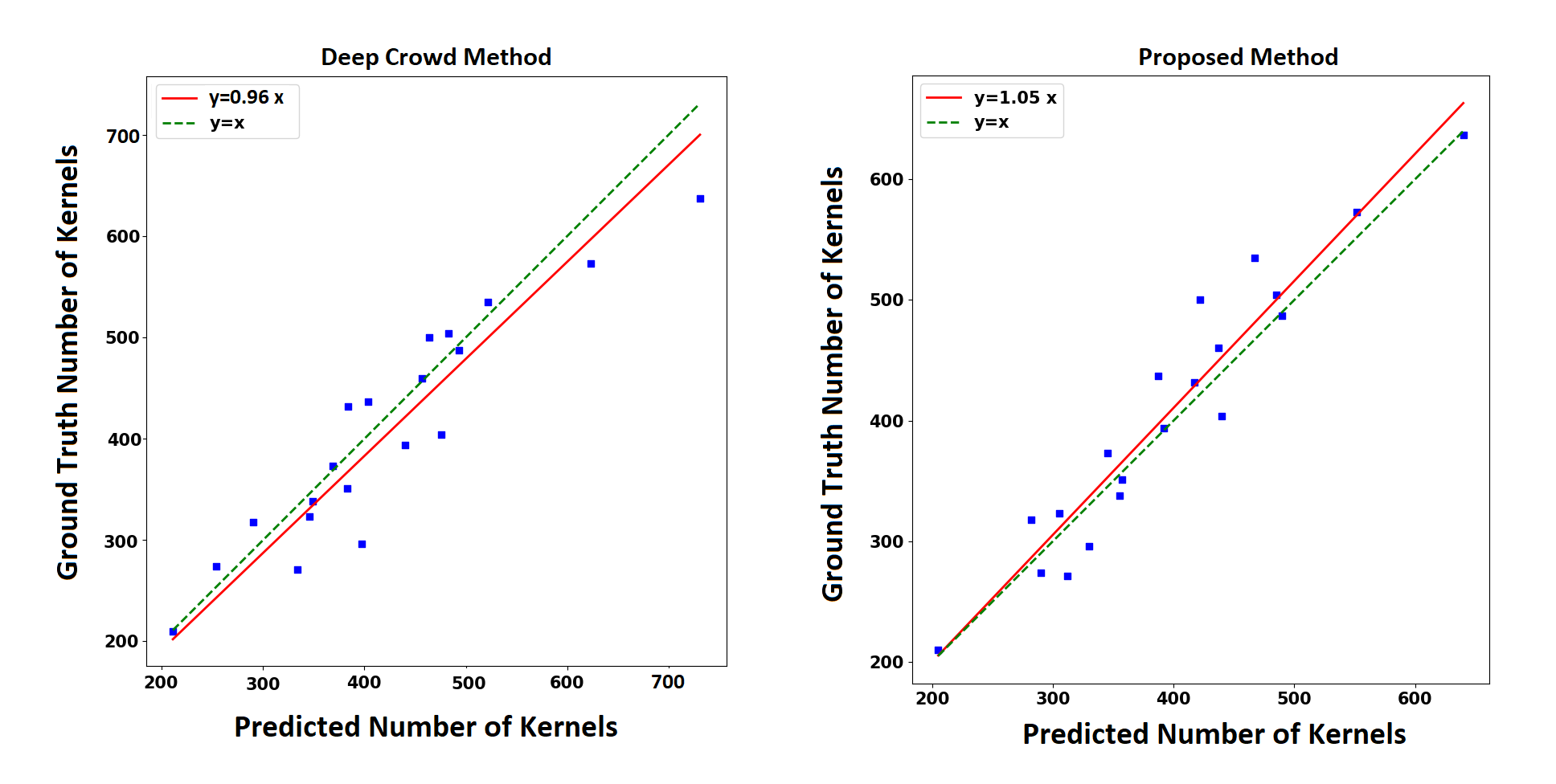}
   \caption{The left and right plots show the predicted number of kernels versus ground truth number of kernels for the Deep Crowd method and proposed method, respectively.}
    \label{fig:yvsy}
\end{figure}

Table \ref{tab:final_res} compares the performances of the competing methods with respect to the root-mean-squared error (RMSE), mean absolute error (MAE), and correlation coefficient. Figure \ref{fig:yvsy} shows the plot of the estimated number of kernels versus the ground truth number of kernels. The proposed method outperforms the Deep Crowd method with respect to all performance measures. Compared to the Deep Crowd method which only performs counting without localization, the proposed method performs both localization and counting. However, the Deep Crowd method has a smaller inference time compared to our proposed method. 

\begin{table}[H]
 \caption{The performances of the competing methods on the kernel counting task of 20 different corn ears.}
    \label{tab:final_res}
    \centering
    \begin{tabular}{|c|c|c|c|}
    \hline
   Method     &  RMSE& MAE& Correlation Coefficient \\
   \hline \hline
   Proposed     & 33.11&25.95& 95.86 \\
   \hline 
    Deep Crowd \cite{wang2015deep}   & 45.29&35.25& 93.12 \\
    \hline
    \end{tabular}
   
\end{table}

\section{Discussion}
In this paper, we propose a kernel detection and counting method based on the sliding window approach. The proposed method detects and counts kernels on single or multiple corn ears from an image. The sliding window approach uses a CNN classifier for kernel detection. Then, a non-maximum suppression is applied to remove overlapping detections. Finally, windows that are classified as kernel are passed to a regression model for finding the $(x,y)$ coordinates of the center of kernel image patches. Due to the effectiveness of the CNN classifier, this approach does not make any assumptions on the lighting conditions, the background quality or the number of ears, or the orientation of the ear like previous approach do. Removing these limitations allows farmers and agronomists to use this in-field to estimate the number of kernels on an ear of corn, given them additional decision making power when it comes to their crop. Moreover, we did not use popular object detection methods such as SSD \cite{liu2016ssd}, YOLO \cite{redmon2016you}, and fast R-CNN \cite{girshick2015fast} mainly because these methods need considerable amount of annotated images which do not exist publicly for the corn kernel detection. In addition, we could not use transfer learning since corn kernel detection is very different than other object detection tasks such as car and human detections and features learned from pre-trained models cannot be easily transferred to our kernel detection task.

When comparing with the other object counting model, our experiments suggest the effectiveness of the proposed method is superior in both the detection and counting of corn kernels. Our proposed method is able to detect kernels on a batch of corn ears at different angles. This approach could be extended to address several future research directions. For example, similar approach could be used for disease detection and quality assessment of corn.
 
%%%%%%%%%%%%%%%%%%%%%%%%%%%%%%%%%%%%%%%%%%
\vspace{6pt} 

%%%%%%%%%%%%%%%%%%%%%%%%%%%%%%%%%%%%%%%%%%
%% optional
%\supplementary{The following are available online at \linksupplementary{s1}, Figure S1: title, Table S1: title, Video S1: title.}


%%%%%%%%%%%%%%%%%%%%%%%%%%%%%%%%%%%%%%%%%%
\authorcontributions{``conceptualization, S.K., H.P., Y.H. A.K., and W.K.; methodology, S.K. and H.P.; software, S.K.; validation, S.K. and H.P.; formal analysis, S.K.; data curation, S.K., H.P., Y.H., A.K., and W.K.; writing--original draft preparation, S.K. and H.P.; writing--review and editing, S.K., H.P., Y.H., A.K., W.K. and L.W.; visualization, S.K.; funding acquisition, L.W.''.}

%%%%%%%%%%%%%%%%%%%%%%%%%%%%%%%%%%%%%%%%%%
\funding{This work was partially supported by the National Science Foundation under the LEAP HI and GOALI programs (grant number 1830478) and under the EAGER program (grant number 1842097). Additionally this work was partially supported by Syngenta.}

%%%%%%%%%%%%%%%%%%%%%%%%%%%%%%%%%%%%%%%%%%
%\acknowledgments{We thank Syngenta and the Analytics Society of INFORMS for organizing the Syngenta Crop Challenge and providing the valuable datasets}

%%%%%%%%%%%%%%%%%%%%%%%%%%%%%%%%%%%%%%%%%%
\conflictsofinterest{The authors declare no conflict of interest.} 

%%%%%%%%%%%%%%%%%%%%%%%%%%%%%%%%%%%%%%%%%%
%% optional

%%%%%%%%%%%%%%%%%%%%%%%%%%%%%%%%%%%%%%%%%%
%% optional
\appendixtitles{no} %Leave argument "no" if all appendix headings stay EMPTY (then no dot is printed after "Appendix A"). If the appendix sections contain a heading then change the argument to "yes".


%=====================================
% References, variant A: internal bibliography
%=====================================
%\reftitle{References}
%\begin{thebibliography}{999}
% Reference 1
%\bibitem[Author1(year)]{ref-journal}
%Author1, T. The title of the cited article. {\em Journal Abbreviation} {\bf 2008}, {\em 10}, 142--149.
% Reference 2
%\bibitem[Author2(year)]{ref-book}
%Author2, L. The title of the cited contribution. In {\em The Book Title}; Editor1, F., Editor2, A., Eds.; Publishing House: City, Country, 2007; pp. 32--58.
%\end{thebibliography}

% The following MDPI journals use author-date citation: Arts, Econometrics, Economies, Genealogy, Humanities, IJFS, JRFM, Laws, Religions, Risks, Social Sciences. For those journals, please follow the formatting guidelines on http://www.mdpi.com/authors/references
% To cite two works by the same author: \citeauthor{ref-journal-1a} (\citeyear{ref-journal-1a}, \citeyear{ref-journal-1b}). This produces: Whittaker (1967, 1975)
% To cite two works by the same author with specific pages: \citeauthor{ref-journal-3a} (\citeyear{ref-journal-3a}, p. 328; \citeyear{ref-journal-3b}, p.475). This produces: Wong (1999, p. 328; 2000, p. 475)

%=====================================
% References, variant B: external bibliography
%=====================================
\externalbibliography{yes}
\bibliography{egbib}

%%%%%%%%%%%%%%%%%%%%%%%%%%%%%%%%%%%%%%%%%%
%% optional
%\sampleavailability{Samples of the compounds ...... are available from the authors.}

%% for journal Sci
%\reviewreports{\\
%Reviewer 1 comments and authors’ response\\
%Reviewer 2 comments and authors’ response\\
%Reviewer 3 comments and authors’ response
%}

%%%%%%%%%%%%%%%%%%%%%%%%%%%%%%%%%%%%%%%%%%
\end{document}